# Fine-Tuning a Large Vision-Language Model for Artwork's Scoring and Critique

Zhehan Zhang,[1] Meihua Qian[1], Li Luo[2], Siyu Huang[2], Chaoyi Zhou[2], Ripon Saha[3],

Xinxin Song[4]

[1]College of Education, Clemson University

[2]School of Computing, Clemson University

[3]Department of Computer Engineering, Arizona State University

[4]Dipartimento di Architettura (DiDA), University of Florence

**Author Note**

Correspondence concerning this article should be addressed to Zhehan Zhang, College of Education, Clemson University. Email: zhehanz@clemson.edu

## Abstract

Assessing artistic creativity is foundational to creativity research and arts education. However, common assessment methods, such as the Torrance Tests of Creative Thinking (Torrance, 1988), are labor-intensive to score when administered at a large scale. While prior machine-learning approaches have demonstrated success in visual creativity scoring (Acar et al., 2025), many existing automated systems rely primarily on image features and provide limited or no explanatory feedback. In this paper, we propose a novel framework for automated human painting creativity assessment by fine-tuning the Large Vision-Language Model (VLM) Qwen2-VL-7B. The model uses a multi-task learning strategy to create a new pattern.

Our dataset consists of 1,000 human-created paintings, randomly scored on a 1–100 scale, spanning diverse themes including family life, nature, and abstract concepts. Each painting is paired with a human-written description of its content or the author's explanation. Two expert raters evaluated the artworks using a rubric comprising five dimensions—originality, color, texture, composition, and content—and provided written critiques. An 80/20 train–test split was employed.

Our methodology introduces a specialized regression head integrated with the VLM's visual encoder output, allowing the model to simultaneously predict numerical scores and generate coherent, rubric-aligned feedback in a single run. By embedding a structured scoring rubric and artwork descriptions into the system prompt, we align the model's textual output with its quantitative predictions. Experiments demonstrate that our model achieved excellent performance, with a Pearson correlation coefficient exceeding 0.97 and a Mean Absolute Error (MAE) of approximately 3.95 on a 100- point scale (meaning prediction accuracy=96.05%).

Qualitative analysis also reveals that the generated feedback exhibits a strong semantic similarity with human experts' feedback (Average Semantic Cosine Similarity=0.798), providing succinct and equitable insights.

This work bridges the gap between computer vision and art assessment, offering a scalable tool to support artistic creativity assessment in the future.

Fine-Tuning a Large Vision-Language Model for Artwork's Scoring and Critique

The problem of K12 teachers overworking has been a long-time pain in many public K12 schools in the country which caused lower quality of teaching. The past research shows time pressure is the main reason for K12 public school teachers that cause "Emotional Exhaustion", which eventually would limit each student's personal feedback (Skaalvik & Skaalvik, 2020). According to the latest National Center for Education Statistics (NCES) Reports (2022), many public elementary school art classes have a Student-Teacher Ratio of 500:1, aggressively more than the National Art Education Association (NAEA) suggested 300:1 (NAEA, 2019), which shows an urgent call for helping those art teachers for grading students' artwork. Given the circumstances, using an AI assistant to help K12 art teachers might be a feasible way to try. The past research (Nazaretsky et al., 2022) shows K12 art teachers are not against AI assistants to help them in terms of teaching and grading, but they want "Explainable AI" not the black box AI in terms of helping them grading. When talking about grading elementary school students' artwork, according to the latest framework of Program for International Student Assessment (PISA), creative thinking in arts class is central to creativity research and arts education (OECD, 2022). Art teachers are encouraged to design classes that can show creativity through student artwork or weekly assignments.

However, it remains difficult to scale artistic creativity in students' artwork because rubric-based scoring and critique writing are labor intensive in teaching practice. Traditional creativity assessment including artistic creativity assessment, mostly relies on two main approaches: Divergent Thinking (DT) Tests (it primarily measures creative potential) and the Consensual Assessment Technique (CAT), which focuses to evaluate the actual creative product.

The psychometric approach treats creativity as a measurable cognitive ability, just like intelligence, with a strong emphasis on divergent thinking—the capacity to come up with multiple solutions to some open-ended problems. The Torrance Tests of Creative Thinking (TTCT) are often seen as the gold standard in the field of creativity assessment (Torrance, 1988), along with simple tasks like the using your imagination to expand all the usage of a certain thing, like someone might list all the possible uses for a brick. Responses get scored on four main points: fluency (how many relevant ideas you produce), flexibility (how many different categories those ideas fall into), originality (how unusual your ideas are compared to others), and elaboration (how much detail you add). In contrast, the Consensual Assessment Technique (CAT), developed by Teresa Amabile (1982), takes a social-psychological view by focusing on the creative product itself rather than the person's cognitive abilities. This method works on the idea that creativity is subjective and should be judged by people who actually know the field. The process is straightforward: experienced judges in the relevant domain independently assess creative works relative to each other, with some fixed standard, and they aren't given specific criteria to follow—they rely on their expert's experience and intuition. What's interesting is that experts tend to agree with each other at surprisingly high rates (with a high inter-rater reliability), which makes this technique a reliable way to assess creative performance.

Nevertheless, within the artistic creativity measures, including previously mentioned divergent thinking (DT) instruments, it often requires labor-heavy human scoring, limiting accessibility in large-scale grading scenarios (Cropley & Marrone, 2022). In recent years, machine learning has opened a new path to automate the grading for human artwork, and understanding how it works comes down to two things: how the computer actually learns and what technology it uses to analyze the art. When it comes to machine learning, computers can take either a supervised or unsupervised learning, just like how students learn either through direct instruction or independent exploration. Supervised learning is the more effective method right now—researchers basically show the computer thousands of drawings that human experts have already graded, and over time, the computer picks up on the patterns between visual features and scores, eventually getting good enough to predict scores for new drawings with impressive accuracy. A model called AuDrA (Patterson et al., 2024), for instance, was trained this way and managed to match human judges with a correlation of over .80. On the other hand, unsupervised learning works differently since the computer doesn't get any "correct" answers to learn from; instead, it measures something called semantic distance by converting images into numerical vectors and figuring out how different a student's drawing is from the original prompt or from what other students have done. In 2024, Acar and his collegues (Acar et al., 2024) have done research based on unsupervised learning using data from TTCT. The catch is that unsupervised methods don't handle complex drawings very well and generally perform worse than supervised approaches when it comes to spotting creativity.

Figure 1

*Compare supervised learning and unsupervised learning in automated creativity assessment*

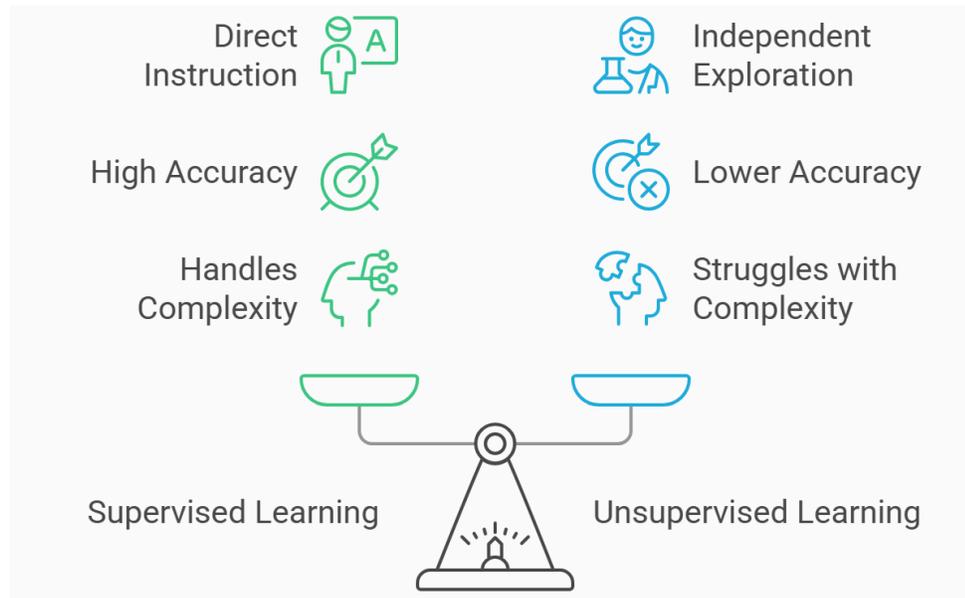

As for the technology itself, different machine learning architectures serve as the system's eyes and brain. Convolutional Neural Networks (CNNs), are built to process image data by scanning artwork for both basic features like edges and textures and more complex ones like shapes and overall composition—researchers have used models like EfficientNet and ResNet to grade paintings based on things like originality, color, and texture, and they've even scored the Test for Creative Thinking-Drawing Production (TCT-DP) with accuracy that rivals human raters (Panfilova et al., 2024). Vision Transformers take a different route by breaking images into small patches and analyzing how those patches relate to each other, similar to how language models work with words in a sentence, and models like BEiT have proven especially good at understanding full images and scoring tests like the Torrance Figural Test. Then there are multimodal models like Contrastive Language-Image Pre-training (CLIP) model that looks at both the image and any text the student provides, which helps when a drawing is hard to interpret on its own—one study found that a model analyzing both a painting and its description hit 95.3% accuracy, way better than image-only models (Zhang et al., 2025). Of course, one big concern is the "black box" problem—how do we know the computer isn't just counting ink blobs?

While automated scoring approaches have improved substantially in recent years. Some educators argue that educational AI tools should interpret the activity—especially when used to generate evaluative judgments or feedback (Nazaretsky et al., 2022). This paper has developed an explainable model, which can not only score each human painting based on given rubric, but also provide score-aligned explanatory feedback for each score rated by the model in five dimensions. We developed a Supervised learning model based on fine-tuning the Qwen2-VL vision-language model (Wang et al., 2024). The training data was presented as a multi-task learning problem, simultaneously optimizing for: 1. Regression Loss: Minimizing the Mean Absolute Error (MAE) between the predicted scores and the ground-truth ratings provided by experts. 2. Generation Loss: Minimizing the cross-entropy loss between the generated feedback and the experts' critiques provided in the dataset.

## Materials and Method

To construct a robust training corpus for automated assessment. Our dataset pulled together 1,000 paintings from three different places to make sure we had a good mix of creative work at different skill levels. We got 750 children's paintings from iStock (with the proper research license) by searching for "kids painting" images. These covered most of the usual subjects that people would expect from kids' art—families, animals, nature scenes, outer space, that kind of thing—and the skill levels were all over the map. We had to work within a budget, so out of roughly 800 available images, we grabbed 750 by picking every other one from the search results. The quality of those children's paintings varies from newbies to sophisticated ones. For the professional artists' paintings, the study went to Wikimedia Commons and collected 200 works from various artists, many of whom are completely unknown. The same approach was used here, just grabbing the first 200 images that met our size requirements. These paintings were generally more polished than the kids' work, with better technique, more detail, and more thought-out compositions. The subjects ranged from portraits to landscapes to abstract ideas, which gave us a lot more visual variety to work with. The study also wanted some truly exceptional examples, so the dataset also added 50 famous masterpieces—think Mona Lisa, Starry Night, pieces like that—which found in public domain collections on Wikimedia Commons or from museums that make their collections available online. These served as our high-creativity benchmarks to see how well the model could recognize outstanding work.

Every painting came with some kind of text description. For the children's and professional paintings, these descriptions either came from the original source, or we wrote them ourselves to capture what the painting showed or what the artist seemed to be going for. Some were straightforward, like "A child's drawing of a family in a house, with smiling stick figures," while others went into more depth about meaning or technique, especially the professional pieces. For the famous works, the descriptions used short art history notes or comments from the artists themselves. To get our ground truth creativity scores, the study brought in two experts who had solid backgrounds in visual arts and creativity research. Each expert looked at all the paintings independently and scored them using a standardized rubric that we'd put together. The average scores for each painting to get a final creativity score out of 100. The two experts agreed with each other almost perfectly—the Intraclass Correlation Coefficient (ICC) came out to 0.99—which told us they were applying the rubric pretty consistently.

The dataset was divided in the usual way: 80% for training and 20% for testing. The study didn't carve out a separate validation set because the dataset wasn't huge to begin with; instead, we let the training process and what we'd learned from previous research guide our decisions about things like hyperparameters, which is the same approach I took in an earlier study. When splitting the data, the process made sure that for every five paintings in each category, one went into the test set, so the data have a good spread of content to test with.

For the expert ratings, the study put together a detailed rubric (see attachment) that scored each painting's creativity across five main dimentions: originality, color, composition, texture and content, building on established frameworks for assessing creativity in art (Amabile, 1982; Torrance, 1988). Each expert rated all five dimensions for each painting and adding them up gave us a total creativity score out of 100. This rubric was based on the Consensual Assessment Technique (CAT) approach—using expert judges—and it also drew criteria that other automated scoring studies have used before. It works for both children's art and professional art. It's worth mentioning that while originality often takes center stage in creativity discussions, we gave all five dimensions equal weight because artistic creativity is really multifaceted—a piece could be wildly original but fall flat in color or composition, or vice versa, and we wanted to capture all of that. The rubric came from earlier work and some trial runs we did beforehand (Zhang et al., 2024). We averaged the two experts' scores for each dimension, and the fact that they agreed so strongly (reached 0.99 ICC) suggests they were on the same page about how to use the rubric.

**Figure 2**

*Model architecture based on Qwen2-VL-7B*

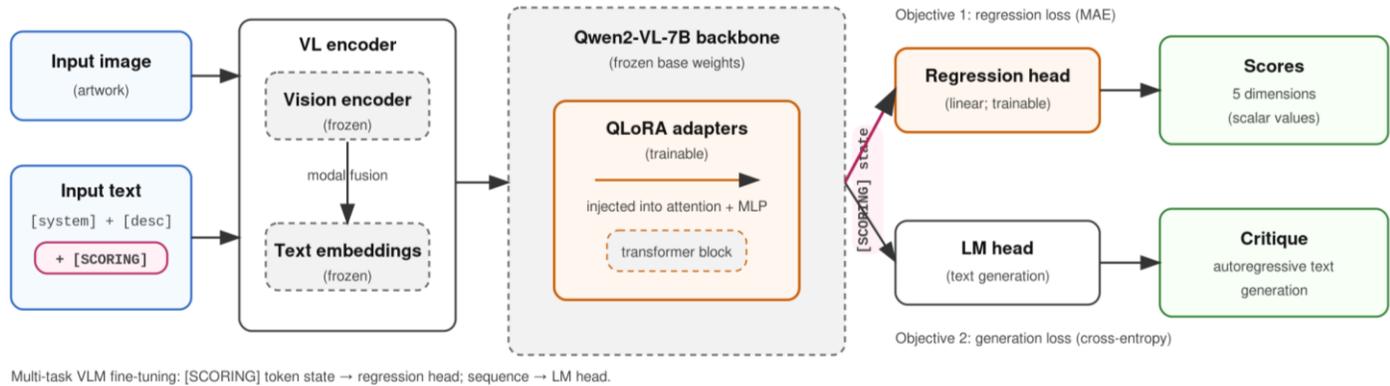

## Model Architecture

The study chose Qwen2-VL-7B-Instruct as the model's backbone architecture. It is more fitting for a vision-language model (VLM) than a traditional convolutional neural network (CNN) to capture the semantic relations between visual elements and aesthetic feedback. Our approach diverges from standard VLM usage by treating assessment as a dual-objective task. We modified the model by attaching a specialized regression head to the final hidden state of the [SCORING] token. This architectural change forces the model to encode visual features into a high-dimensional vector that simultaneously informs a scalar prediction (the score) and an autoregressive text generation sequence (the critique).

## Fine-Tuning

Given the budget limits in the research settings, the study utilized QLoRA (Quantized Low-Rank Adaptation) for training (Dettmers et al., 2024). Through the approach of freezing the pre-trained weights and injecting trainable low-rank adapters into the attention and MLP layers, the model fine-tuning is efficient on the H100 GPU hardware without sacrificing the performance. As for the training loss function, it was designed to decrease deviations in both numerical accuracy (L1 Loss) and textual coherence (Cross-Entropy Loss). This approach ensured the model wound not hallucinate feedback but grounds it in the predicted score.

**Teachers' Feedback Learning**

The model developed a system prompt that embeds the assessment rubric. This prompt was summarized from teachers real feedback and would direct the model to learn those adjectives (e.g., "fair," "exceptional") that align with specific score ranges. To avoid the "black box" of AI grading, we fed the artwork's textual description with the image pixels proved critical in reducing visual hallucinations, allowing the model to correctly interpret ambiguous elements in children's drawings.

## Results

**Predictive Accuracy**

Training stability was observed early in the process, with validation metrics plateauing near Epoch 5. On the test set (N=200), the model demonstrated a high consistency with the human expert ratings.

**Figure 3**

*Human score vs the model prediction for each painting in test dataset*

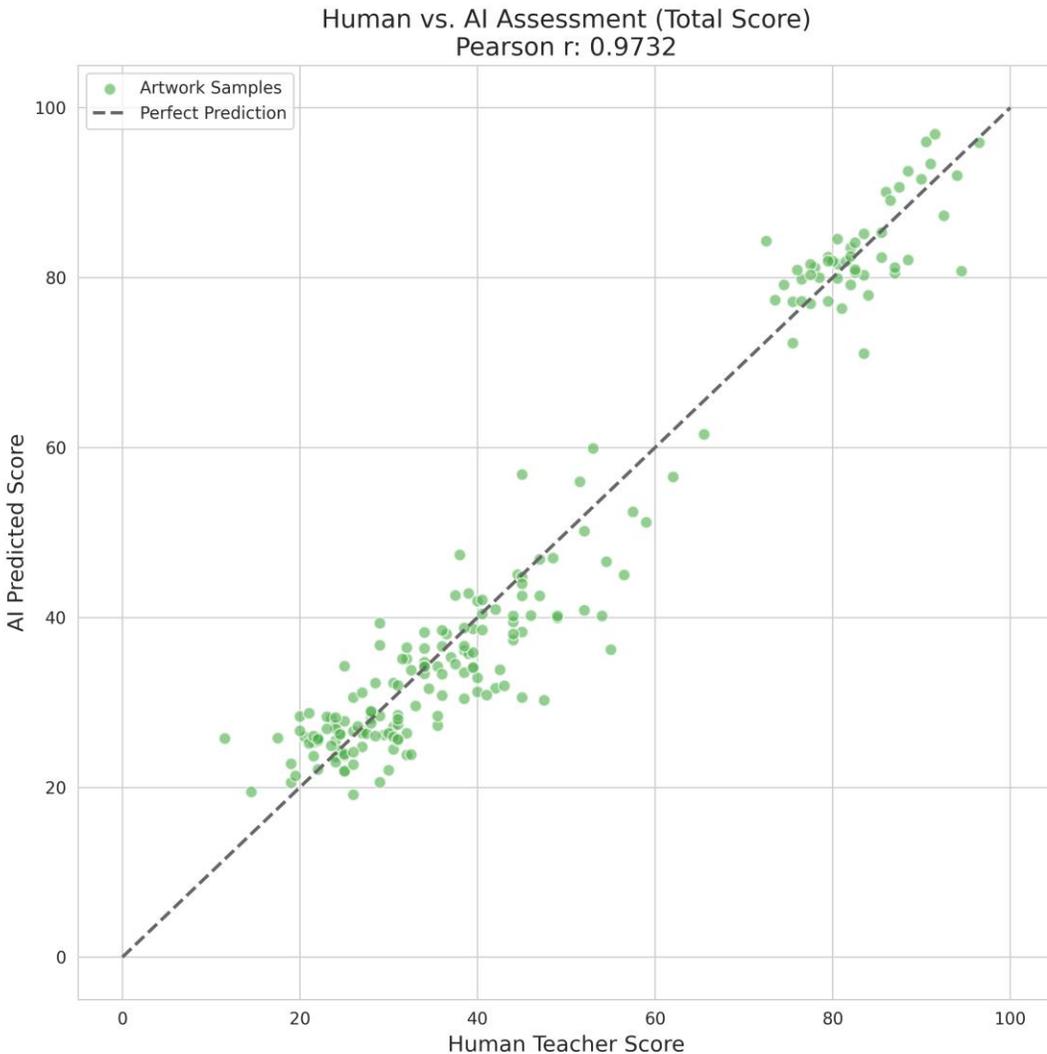

In Figure 3, the model achieved a Pearson correlation coefficient R=0.975 for the Total

Creativity Score. It suggests a near-linear relationship between the AI's predictions and human

judgment. In Figure 4, the Mean Absolute Error (MAE) achieved at approximately 3.95 on a

100-point scale. Statistically, this implies that the AI's grading deviates from a human expert by less than 4 points on average—a margin often indistinguishable from the variance found between two human raters.

**Figure 4**

*Model's training loss and MAE*

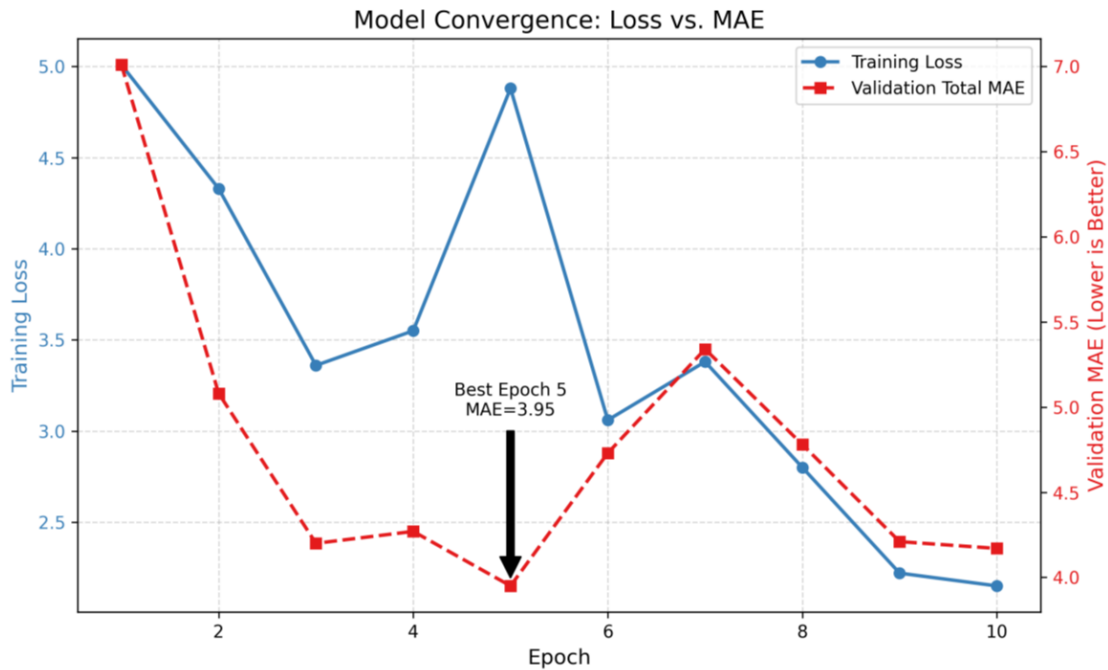

## Quality of Generated Feedback

Beyond the numbers, the semantic quality of the automated critiques was evaluated using Sentence-BERT (SBERT) (Reimers & Gurevych, 2019). This measurement captures the conceptual overlap between machine-generated text and expert's original comments, rather than mere keyword matching. In Figure 3, the average Semantic Cosine Similarity score of the model was 0.798. In computer science terms, a score close to 0.80 means that the generated paragraph is not only producing their own unique ideas on those paintings, but also not copying the experts' words, otherwise the Semantic Cosine Similarity will be more than 0.85, close to 0.9.

**Figure 5**

*Average semantic Cosine Similarity between experts' feedback and model generated feedback*

```
=== Starting Comment Semantic Similarity Evaluation ===
Loading raw SBERT model: sentence-transformers/all-MiniLM-L6-v2...
Comparing comments for 200 test samples...
Analyzing: 100%|████████████| 200/200 [16:50<00:00,  5.05s/it]

==================================================
Average Semantic Cosine Similarity: 0.7980
==================================================
```

## Discussion

This study challenges the notion that creative assessment requires solely human intuition. Leveraging a supervised fine-tuning approach with rubric-aligned VLM indicating that computational models can effectively combine aesthetic criteria like originality and art skills. Unlike unsupervised methods that rely on data feature distances, our supervised multi-task model can learn the language of art critique directly from experts' data. It arguebaly provide a more interpretable link between the visual input and the assigned score. For K-12 art educators facing overwhelming student-teacher ratios, this tool offers a pragmatic solution to the "feedback gap." The high correlation and semantic alignment imply that the model can function as a trusted "second reader", thus automating the laborious practice of initial grading and constructng feedback. This doesn't replace the teacher, but instead it increases their capacity and allows for time spent doing more purposeful, high-level mentorship.

Limitation

It is worth noting that our model's performance relies heavily on the quality of the training data. The study uses a relatively small sample of human paintings as the dataset for model training., which can cause potential bias on certain types of human paintings. For example, the children's paintings in the dataset may not fully capture the cultural nuances of student art globally, because most of them are from English speaking countries due to the nature of the websites Istock. Future experiments should include more varied training data and check the model's capability to identify differences in cultural context.

Conclusion

We have developed a computer program that grades artworks automatically. It is a rubric-based multi-task framework for the automated assessment of K12 students' artwork. The ability to predict scores has been achieved by fine-tuning the Qwen2-VL model. While the model generate humanized critiques in the same time, we bridged the gap between quantitative scoring and qualitative feedback. With a prediction of accuracy exceeding 96% and highly consistent textual outputs, this system represents a significant step toward scalable, equitable, and explainable AI in arts education.